\documentclass[10pt,twocolumn,letterpaper]{article}
\usepackage[pagenumbers]{cvpr}
\pdfoutput=1

\usepackage{graphicx}
\usepackage{amsmath}
\usepackage{amssymb}
\usepackage{booktabs}
\usepackage{tabularray}
\usepackage{array}
\usepackage{booktabs}
\usepackage{tabularray}
\usepackage{rotating}
\usepackage{textcomp, gensymb}

\usepackage[pagebackref,breaklinks,colorlinks]{hyperref}

\usepackage[capitalize]{cleveref}
\crefname{section}{Sec.}{Secs.}
\Crefname{section}{Section}{Sections}
\Crefname{table}{Table}{Tables}
\crefname{table}{Tab.}{Tabs.}

\def\cvprPaperID{10085} % *** Enter the CVPR Paper ID here
\def\confName{CVPR}
\def\confYear{2023}

\newcommand{\AD}[1]{\textcolor{blue}{}}

\newcommand{\TK}[1]{\textcolor{red}{}}

\begin{document}

\title{ORCa: Glossy Objects as Radiance-Field Cameras}
% Other title ideas:
% ROAR: Reflective Objects as Radiance Fields 
% ROAR: Reflective Objects as Radiance Fields Cameras
% RACRO: Radiance-Fields Cameras from Reflective Objects
% ROC: Reflective Objects as Cameras
%
\newcommand{\superscript}[1]{\ensuremath{^{\textrm{#1}}}}
\author{
    Kushagra Tiwary\superscript{1\textbf{*}}, Akshat Dave\superscript{2\textbf{*}},
    Nikhil Behari\superscript{1},
    Tzofi Klinghoffer\superscript{1},\\
    Ashok Veeraraghavan\superscript{2},
    Ramesh Raskar\superscript{1}\\
    \superscript{1}Massachusetts Institute of Technology, \superscript{2}Rice University\\
    {\tt\small \{ktiwary,behari,tzofi,raskar\}@mit.edu, \{akshat,vashok\}@rice.edu}
    %\vspace{-20pt} % reduce spacing before figure
}

%\and
%%%%% Teaser Figure
\twocolumn[{%
\renewcommand\twocolumn[1][]{#1}%
\maketitle
\begin{center}
    \centering
    \captionsetup{type=figure}
    \label{fig:teaser}
    \includegraphics[width=0.9\textwidth]{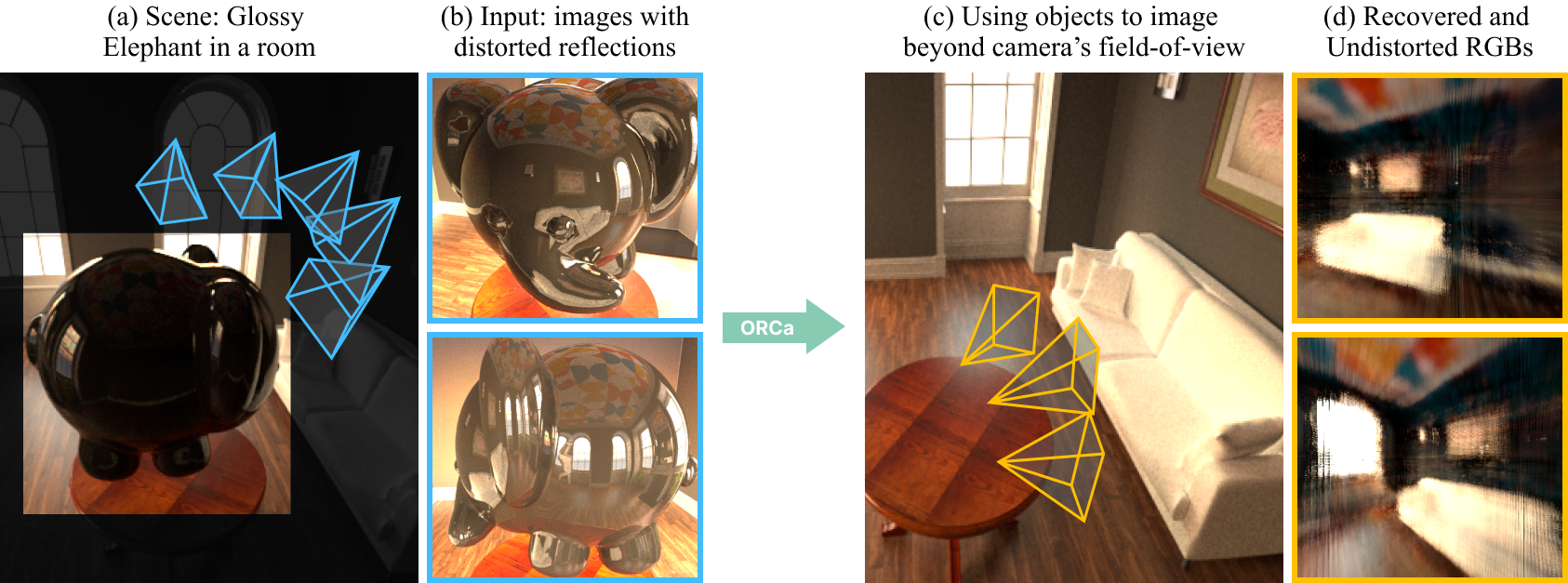}
    \captionof{figure}{\textbf{Objects as radiance-field cameras}.
    We convert everyday objects with unknown geometry (a) into radiance-field cameras by modeling multi-view reflections (b) as projections of the 5D radiance field of the environment. We convert the object surface into a virtual sensor to capture this radiance field (c), which enables depth and radiance estimation of the surrounding environment. We can then query this radiance field to perform beyond field-of-view novel view synthesis of the environment (d).}
    %   from the object to its surroundings in addition to being  occlusion aware for close-by objects. We also show that we can query this radiance field to render views that are otherwise only visible to the object and beyond the field-of-view of the real camera.
    % From multi-view images of a glossy object of unkown geometry  we recover radiance fields of surrounding environment beyond camera's field of view (a), by modeling the object surface as a virtual sensor (b). }
\end{center}%
}]

\maketitle

%%%%%%%%% ABSTRACT
\begin{abstract}
\noindent 
Reflections on glossy objects contain valuable and hidden information about the surrounding environment. By converting these objects into cameras, we can unlock exciting applications, including imaging beyond the camera's field-of-view and from seemingly impossible vantage points, e.g. from reflections on the human eye. However, this task is challenging because reflections depend jointly on object geometry, material properties, the 3D environment, and the observer viewing direction. Our approach converts glossy objects with unknown geometry into radiance-field cameras to image the world from the object's perspective. Our key insight is to convert the object surface into a virtual sensor that captures cast reflections as a 2D projection of the 5D environment radiance field visible to the object. We show that recovering the environment radiance fields enables depth and radiance estimation from the object to its surroundings in addition to beyond \textit{field-of-view} novel-view synthesis, i.e. rendering of novel views that are only directly-visible to the glossy object present in the scene, but not the observer. Moreover, using the radiance field we can image around occluders caused by close-by objects in the scene. Our method is trained end-to-end on multi-view images of the object and jointly estimates object geometry, diffuse radiance, and the 5D environment radiance field. For more information, visit our  \href{https://ktiwary2.github.io/objectsascam/}{website}. 

\end{abstract}

\vspace{-5mm}
%%%%%%%%% BODY TEXT
\section{Introduction}
\begin{figure*}
    \centering
    \includegraphics[width=\textwidth]{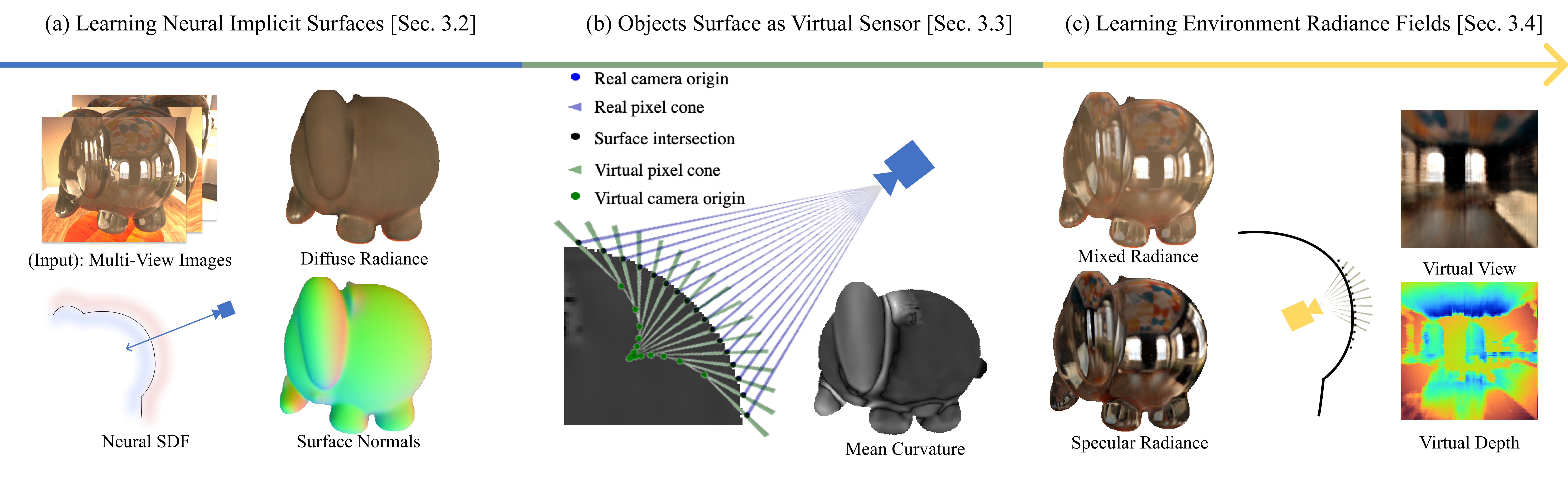}
        \caption{\textbf{ORCa Overview}. We jointly estimate the glossy object's geometry and diffuse along with the environment radiance field estimation through a three-step approach. First, we model the object as a neural implicit surface (a). We model the reflections as probing the environment on virtual viewpoints (b) estimated analytically from surface properties. We model the environment as a radiance field queried on these viewpoints (c). Both neural implicit surface and environment radiance field are trained jointly on multi-view images of the object using a photometric loss.}
    \label{fig:overview}
\end{figure*}

% use reflections cues infer information about what's around the blind spots  
% Why is this problem difficult: 
% What 

% Parked car provide informtion about invisible objects, we are able to do the decomposition of the reflected object and the reflected image. 

% WHat's the problem? WHy is it interesting? Why is it challenging? 
Imagine that you're driving down a city street that is packed with lines of parked cars on both sides. Inspection of the cars' glass windshields, glossy paint and plastic reveals sharp, but faint and distorted views of the surroundings that might be otherwise hidden from you. Humans can infer depth and semantic cues about the occluded areas in the environment by processing reflections visible on reflective objects, internally decomposing the object geometry and radiance from the specular radiance being reflected onto it. Our aim is to decompose the object from its reflections to "see" the world from the object's perspective, effectively turning the object into a camera which images its environment. However, reflections pose a long-standing challenge in computer vision as the reflections are a 2D projection of an unknown 3D environment that is distorted based on the shape of the reflector. 

% Undistorting these reflections to uncover the 3D world will allow for applications that can image beyond the observer's field-of-view, or even from seemingly impossible vantage points such as the human eye. 

% Imagine driving through a street of parked cars. Each parked car reflects light onto the driver's eyes illuminating areas that may otherwise be invisible to the driver. Reflections are ubiquitous and often encode hidden information not directly in direct-line-of-sight of the camera. 

\begin{figure}
    \centering
\includegraphics[width=\columnwidth]{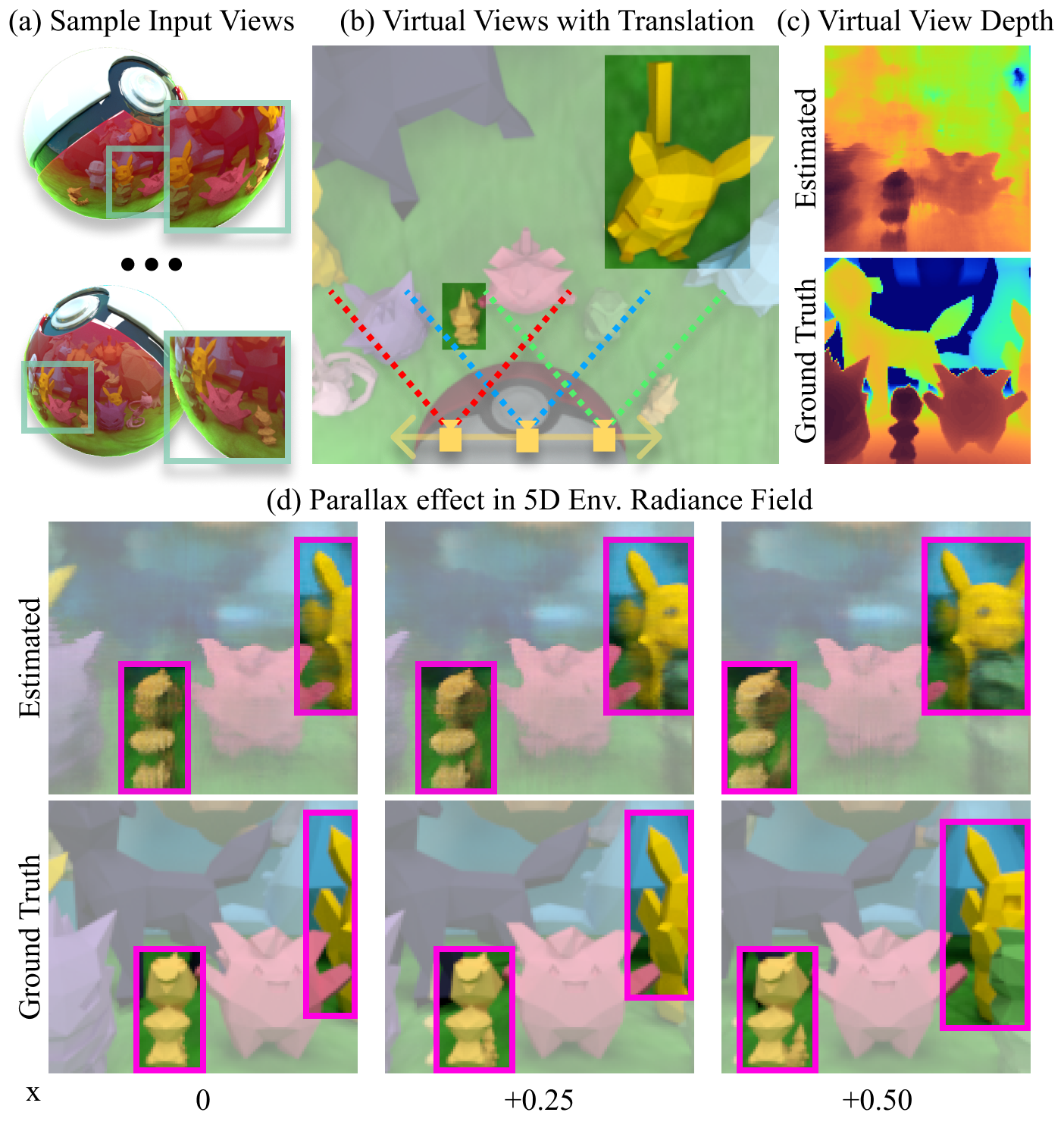}
    \caption{\textbf{Advantages of 5D environment radiance field}. Modeling reflections on object surfaces (a) as a 5D env. radiance field enables beyond \textit{field-of-view} novel-view synthesis, including rendering of the environment from translated virtual camera views (b). Depth (c) and environment radiance of translated and parallax views can further enable imaging behind occluders, for example revealing the tails behind the primary Pokemon occluders (d).}
    \label{fig:parallax_example}
\end{figure}

To capture the 3D world from the object's perspective, we model the object's surface as a virtual sensor that captures the 2D projection of a 5D environment radiance field surrounding the object. This environment radiance field consists largely of areas only visible to the observer through the object's reflections. Our use of environment radiance fields not only enables depth and radiance estimation from the object to its surroundings, but also enables beyond \textit{field-of-view} novel-view synthesis, i.e. rendering of novel views that are only directly visible to the glossy object present in the scene, but not the observer. Unlike conventional approaches that model the environment as a 2D map, our approach models it as a 5D field without assuming the scene is infinitely far away. Moreover, by sampling the 5D radiance field, instead of a 2D map, we can capture depth and images around occluders, such as close-by objects in the scene, as shown in Fig. \ref{fig:parallax_example}. These applications cannot be done from a 2D environment map. 
%\TK{suggest making the next clause its own sentence or rewriting this sentence to be less long}and models position and the viewing direction to the environment from the object.
% , our recovered environment field is dependent on not just the viewing direction but also the 3D point thus handling occlusions caused due by close-by objects in the scene.

We aim to decompose reflections on the object's surface, from its surface and exploit those reflections to construct a radiance field surrounding the object, therefore capturing the 3D world in the process. This is a challenging task because the reflections are extremely sensitive to local object geometry, viewing direction and inter-reflections due to the object's surface. To capture this radiance field, we convert glossy objects with unknown geometry and texture into radiance-field cameras. Specifically, we exploit neural rendering to estimate the local surface of the object viewed from each pixel of the real-camera. We then convert this local surface into a virtual pixel that captures radiance from the environment. This virtual pixel captures the environment radiance as shown in Fig \ref{fig:virtual_cone_with_surface}. We estimate the outgoing frustum from the virtual pixel as a cone that samples the scene. By sampling the scene from many virtual pixels on the object surface, we construct an environment radiance field that can be queried independently of the object surface, enabling beyond \textit{field-of-view} novel-view synthesis from previously unsampled viewpoints.

Our approach jointly estimates object geometry, diffuse radiance, and the environment radiance field from multi-view images of glossy objects with unknown geometry and diffuse texture in three steps. First, we use neural signed distance functions (SDF) and an MLP to model the glossy object's geometry as a neural implicit surface and diffuse radiance, respectively, similar to PANDORA \cite{dave2022pandora}. Then, for every pixel on the observer's camera, we estimate the virtual pixels on the object's surface based on the estimated local geometry from the neural SDF. We analytically compute parameters of the virtual cone through the virtual pixel. Lastly, we use the cone formulation in MipNeRF \cite{barron2021mipnerf} to cast virtual cones from the virtual camera to recover the environment radiance \AD{mention mask}. 

% \paragraph{Contributions} 
To summarize, we make the following contributions: 
\begin{itemize}
    % converting implicit surfaces into virtual viewpoints 
    \item We present a method to convert implicit surfaces into virtual sensors that can image their surroundings using virtual cones. (Sec. \ref{sec:Virtual_Cones})
    \item We jointly estimate object geometry, diffuse radiance, and estimate the 5D environment radiance field surrounding the object. (Fig. \ref{fig:renddataimresults} \& \ref{fig:realdataimresults})
    \item We show that the environment radiance field can be queried to perform \textit{beyond-field-of-view} novel viewpoint synthesis, i.e render views only visible to the object in the scene (Section \ref{sec:env_rad_fields})
\end{itemize}
% \paragraph{Turning objects into cameras.} 

\textbf{Scope.} We only model glossy objects with low roughness as such specular reflections tend to have a low signal-to-noise ratio, therefore are a blurrier estimate of environment radiance field. However, we note that the virtual cone computation can be extended to model the cone radius as a function of surface roughness. Deblurring approaches can further improve resolution of estimated environment. In addition, we approximate the local curvature using mean curvature, which fails for objects with varying radius of curvature along the tangent space. We explain how our virtual cone curvature estimation can be extended to handle general shape operators in the supplementary material. Lastly, similar to other multi-view approaches, our approach relies on a sufficient virtual baseline between virtual viewpoints to recover the environment radiance field.

\section{Related Work}
\subsection{Modeling reflections}
% \subsection{Imaging with multiple views}
\noindent
Catadioptric imaging systems aim to expand the field of view of conventional cameras using reflective mirrors \cite{theory_of_catd_imagin1998} \cite{nayar1997catadioptric}. Recent work in catadioptric imaging proposes using ellipsoidal mirrors to increase the baseline of a camera, such that more of the light is observed \cite{de2022wide} and novel view synthesis from a single capture \cite{wang2021mirrornerf}. These works assume the geometry of the reflecting surface is known or calibrated. In contrast to these methods, we create a catadioptric imaging system from everyday glossy objects of unknown geometry.
\vspace{2mm}

Recent progress in neural radiance fields (NeRF) has enabled impressive novel view rendering and geometry reconstruction from multi-view images \cite{mildenhall2021nerf}. NeRF does this by sampling the 5D light field of the scene and learning a representation that is consistent with the training images. MipNeRF\cite{barron2021mipnerf} demonstrates better novel view synthesis by modeling outgoing rays as cones to enable anti-aliasing. However MipNeRF fails to model sharp view dependencies of reflections. RefNeRF \cite{verbin2022refnerf} shows improved novel view synthesis on reflections using Integrated-Directional Encoding to explicitly separate diffuse and specular radiance. Similarly, NeRFRN \cite{Guo_2022_CVPR} separates diffuse and specular radiance by using separate neural networks. Neural Catacaustics \cite{kopanas2022neural} propose a neural warping method to improve novel view synthesis of reflections by learning the caustics of the surface. Comparatively, while all such works improve the quality of novel-view synthesis from the scene to the primary camera, we perform view synthesis that is beyond the line-of-sight of the primary camera, i.e. rendering views only visible to the objects present in the scene, while jointly estimating object geometry and separating diffuse and specular radiance. We perform \emph{beyond line-of-sight} view synthesis by extracting a 5D environment radiance field from the target object. 

\begin{figure}
    \centering
    \includegraphics[width=\columnwidth]{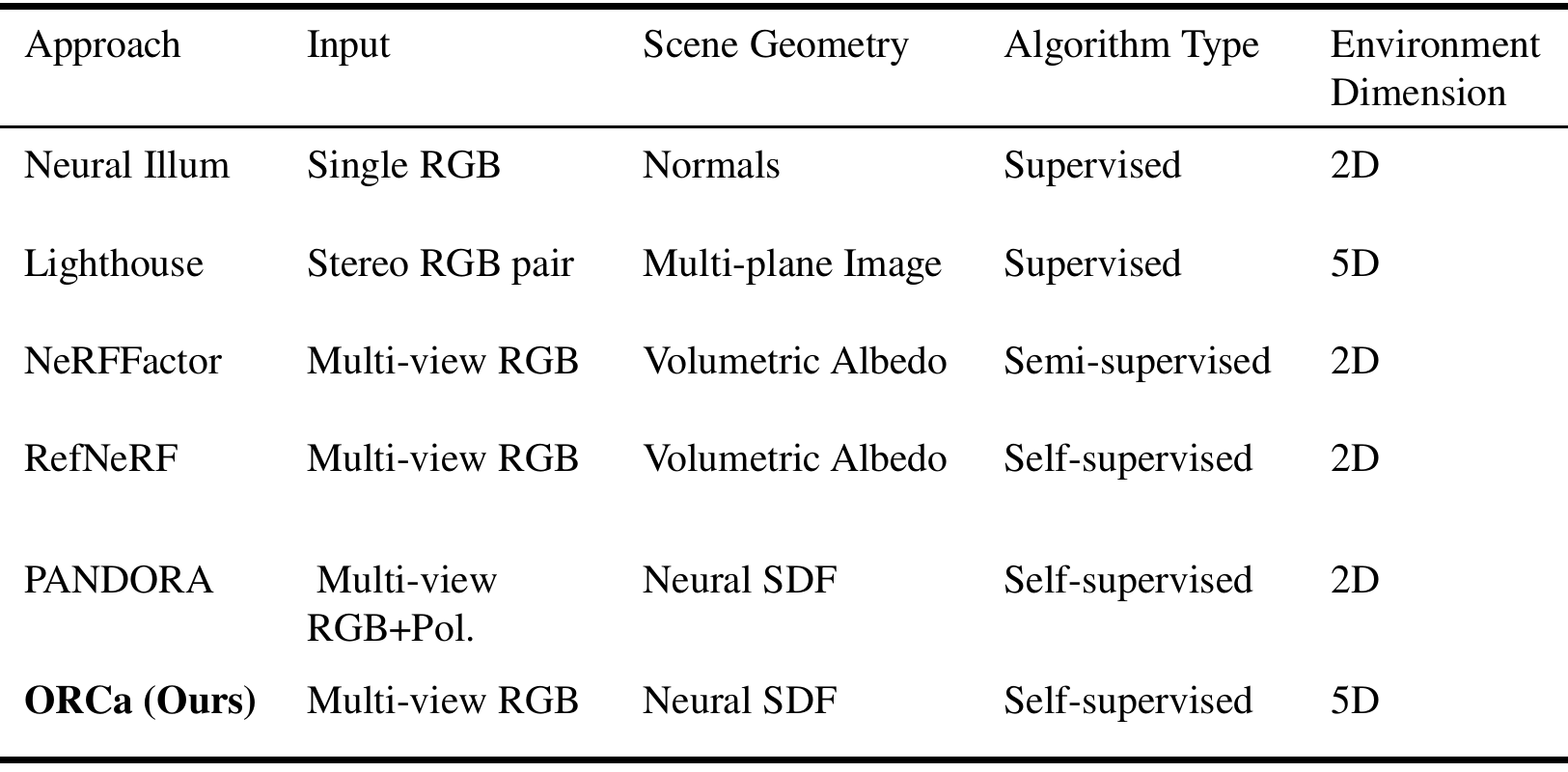}
        \caption{\textbf{Comparison of environment estimation approaches}. 
        Approaches such Neural Illum \cite{song2019neural}, Lighthouse \cite{srinivasan2020lighthouse}, NeRFFactor \cite{zhang2021nerfactor} train on datasets of natural illumination maps to regularize ill-posedness of environment estimation. PANDORA \cite{dave2022pandora} and RefNeRF\cite{verbin2022refnerf} exploit multi-view reflections on object but approximate surrounding environment is infinitely far away and model it with a flat 2D map. From multi-view reflections, ORCa converts the object surface into virtual sensor and extracts 5D radiance field of the environment.\AD{Table in powerpoint}}
    \label{fig:comparisons}
\end{figure}

\subsection{Environment Estimation}
Recovering underlying scene properties from multiple images is inherently ill-posed \cite{ramamoorthi2001signal}, but can be regularized using the natural statistics of scene properties as a prior \cite{romeiro2010blind}\cite{barron2014shape}. Recent works exploit this prior through deep neural networks and demonstrate inverse rendering of indoor scenes from a single image \cite{garon2019fast} \cite{li2020inverse} \cite{wang2021learning} \cite{zhu2022irisformer}. However, these techniques typically recover only coarse representations of lighting and cannot reconstruct fine details of the environment. Lombardi \textit{et al.} \cite{lombardi2012reflectance} recover environment and reflectance, assuming the scene is composed of known geometry and uniform material. Georgolis \textit{et al.} \cite{georgoulis2017around} recover the environment map behind the camera from a single image of a glossy object, assuming the object is composed of textureless materials and using ground truth segmentation masks. Song \textit{et al.}\cite{song2019neural} estimate plausible environment maps by mapping reflections in the image and inpainting unmapped regions. Srinivasan \textit{et al.} \cite{srinivasan2020lighthouse} capture stereo image pairs and estimate plausible spatially-coherent environment maps. NeRD \cite{boss2021nerd}, NeRFactor \cite{zhang2021nerfactor} and NeuralPIL \cite{boss2021neural} employ data-driven priors for lighting and BRDF in a NeRF-based approach for radiance decomposition from multi-view images. 

While the above approaches, which rely on scene priors, can generate realistic environment maps suitable for virtual object insertion and re-lighting, the actual environment might consist of occlusions. Other imaging modalities and properties of light can aid in extracting information about the surrounding environment. Park \textit{et al.} \cite{park2020seeing} use RGB-D videos to estimate environment map. Swedish \textit{et al.} \cite{swedish2021objects} recover high-frequency illumination map from the shadows of an object with known geometry. PhySG \cite{zhang2021physg} and Munkberg \textit{et al.} \cite{munkberg2022extracting} perform inverse rendering from multi-view images by modeling the surface as signed distance functions. PANDORA \cite{dave2022pandora} performs radiance decomposition from polarized RGB images. \AD{mention NeIF}

% Our work builds on top of \cite{volsdf} and \cite{pandora} enabling joint estimation of object geometry, diffuse and specular radiance however. We are also inspired by \cite{mipnerf} and sample virtual cones from the object. We note that by doing this we also improve novel view synthesis quality from the primary camera as well. 

% \begin{figure}
%     \centering
%     \includegraphics[width=0.5\textwidth]{figures/key_idea.pdf}
%     \caption{\textbf{Virtual Environment Radiance Fields}}
%     \label{fig:key_idea}
% \end{figure}
% \begin{figure}
%     \centering
%     \includegraphics[width=0.5\textwidth]{figures/virtual_cone_with_surface.pdf}
%     \caption{\textbf{Effect of surface curvature on virtual cones}}
%     \label{fig:virtual_cone_with_surface}
% \end{figure}
% \begin{figure}
%     \centering
%     \includegraphics[width=0.5\textwidth]{figures/catacaustic_sphere.pdf}
%     \caption{\textbf{Virtual cone computation for sphere}}
%     \label{fig:virtual_cone_with_surface}
% \end{figure}

\section{Learning environment radiance fields from multi-view reflections}

\begin{figure}
    \centering
    \includegraphics[width=0.4\textwidth]{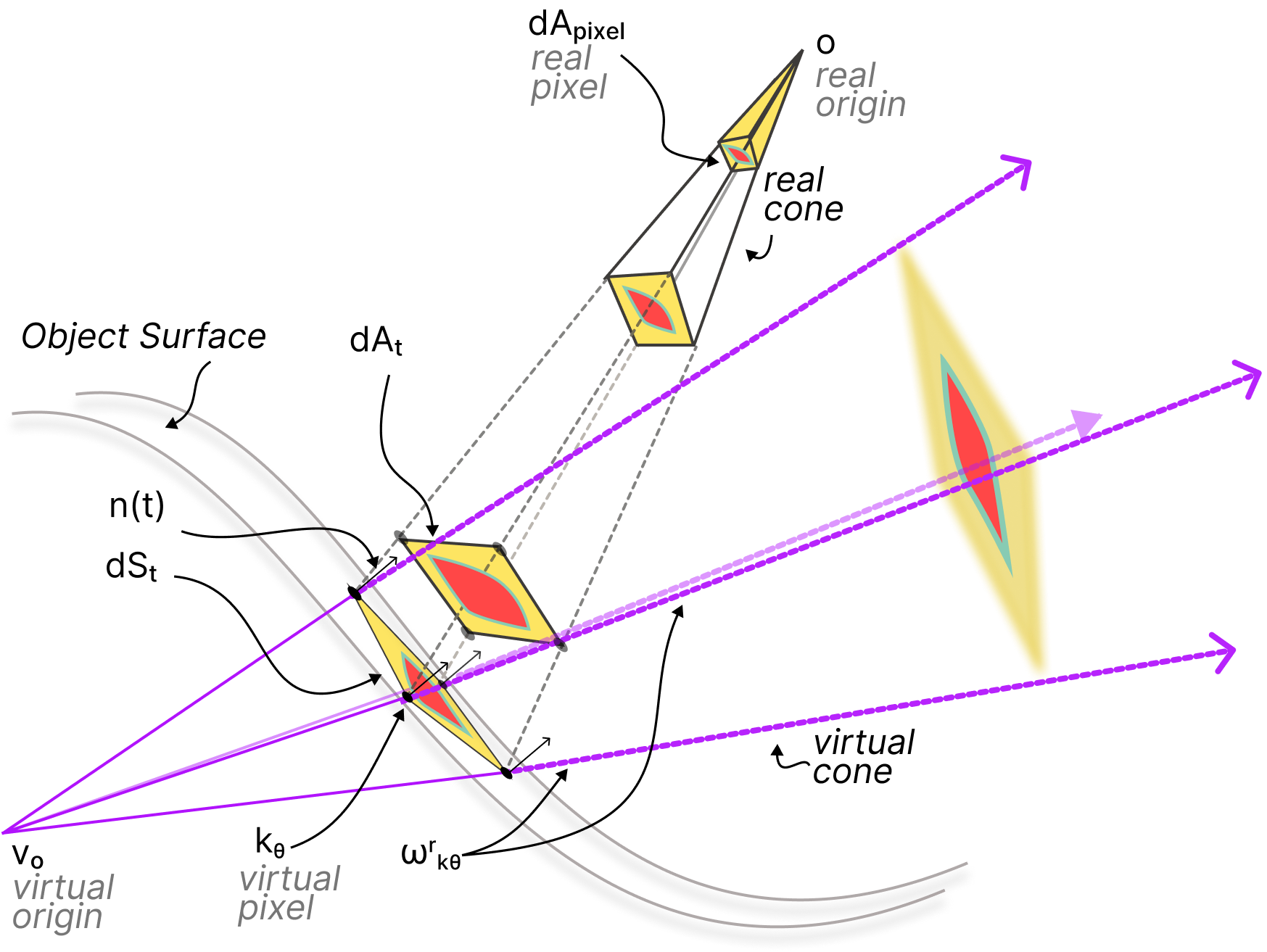}
    \caption{\textbf{Virtual Sensor}. We image the world through the object by modeling each pixel's specular radiance as a projection of the 5D radiance field of the environment onto the object's surface. We capture the radiance field by treating the surface area on the object that the pixel views, $\mathbf{dS}_t$, as a single-pixel virtual camera with its center-of-projection at $\mathbf{v}_o$. We cast virtual cones through the virtual sensor to capture the 5D radiance field of the environment.}
    \label{fig:virtual_cone_with_surface}
\end{figure}

% Our goal is to convert a glossy object with unknown geometry, material properties and albedo into a virtual camera to render the scene's radiance and depth from the object's perspective. In essence, our goal is to "see" the scene from the object's perspective. In the following section we discuss our proposed pipeline. In Section \ref{sec:NIS}, we discuss our approach to learn object properties as neural implicit surfaces, which enables the joint estimation of object geometry and diffuse radiance. We then discuss our formulation to estimate the differential pixel-object intersection surface area, which forms a virtual sensor in Section \ref{sec:Virtual_Cones}. We then present a method to estimate the virtual center of projection for this virtual sensor. We empirically show that our virtual center-of-projection lies on the caustic curve of the object in \ref{sec:virtual_vp}. Finally, in Section \ref{sec:env_rad_fields} we show how we can learn 5D environment radiance fields by sampling single-pixel virtual sensors on the object by shooting virtual cones. With our learned environment radiance fields, we can render novel viewpoints otherwise not in the field-of-view of the original camera by placing a virtual camera on the object.
\subsection{Overview}
 Reflections on glossy objects offer a glimpse into the surrounding environment beyond the camera's field-of-view. From multi-view images of a glossy object with unknown geometry and albedo, we aim to recover the 5D radiance field of the surrounding environment. The mapping from images captured by the observer to the surrounding environment depends on the glossy object's surface properties, in particular, the surface normals and curvature. We first cast a cone from the observer camera's center-of-projection through each pixel viewing the scene. When the cone intersects the object surface, it reflects, causing the cone to be transformed (Fig. \ref{fig:virtual_cone_with_surface}). The transformed cone, referred to as a virtual cone, samples the environment and is primarily responsible for the specular radiance observed on the glossy object. Our key insight is that the reflections captured by the observer's camera can be modeled as a projection of the environment radiance field on to the object surface. By modeling the reflected rays as a cone and computing the parameters of the cone, we can more accurately estimate the projected environment radiance field onto the pbject surface, as shown in Fig. \ref{fig:ivc_vs_cvc}. 
 
ORCa is composed of three steps: modeling the object's geometry as a neural implicit surface (Sec. \ref{sec:NIS}), converting the object's surface into a virtual sensor (Sec. \ref{sec:Virtual_Cones}), and modeling the environment radiance field as a projection along these virtual cones (Sec. \ref{sec:env_rad_fields}). The learned environment radiance field can then be queried on novel viewpoints to show occluded areas in the scene. Fig. \ref{fig:overview} depicts our output for each component on a scene rendered with a complex glossy object and 3D environment. Fig. \ref{fig:architecture} shows our system architecture. Next, we describe each step in detail.   

% \subsection{Preliminaries}
% \begin{itemize}
%     \item NeRF formulation
% \end{itemize}

\subsection{Learning Neural implicit Surfaces}
\label{sec:NIS}

% \textbf{Neural Radiance Fields.} 
\noindent 
% Our method does not assume information about object geometry, therefore we first estimate object geometry and diffuse radiance by representing objects as neural implicit surfaces. We represent the target object using signed distance fields (SDFs) represented by the function $f_{\mathcal{S}}: \mathbb{R}^3 \rightarrow \mathbb{R}$, which maps spatial coordinates in the 3D scene, $\mathbf{x} \in \mathbb{R}^3$, into the signed distance from the camera origin to the object. Our surface, $\mathcal{S}$, is then represented by the zero-level set of the SDF: 
% \begin{equation}
\textbf{Neural Signed Distance Function} We model the object geometry as a neural signed distance function (SDF). 
$f: \mathbb{R}^3 \rightarrow \mathbb{R}$. SDFs provide a helpful inductive bias for learning smooth surface geometry \cite{yariv2021volume}\cite{wang2021neus}\cite{oechsle2021unisurf} that assists  downstream tasks in our pipeline. Moreover, the surface properties crucial for our framework, surface normals and curvature, can be conveniently computed from SDFs in a differentiable manner.
Consider the 3D spatial coordinates, $\mathbf{x}$, in the scene. The glossy object surface, $\mathcal{S}$ is then represented by the zero-level set of the SDF
\begin{align} 
% \begin{gather*} 
\label{eq:zero_level_set}
\mathcal{S} = \{f_{\mathcal{S}}(\mathbf{x}) = 0 | \mathbf{x} \in \mathbb{R}^3 \}
\end{align}
Similar to Yariv \textit{et al.} \cite{yariv2021volume}, we model the SDF $f_{\mathcal{S}}$ as a coordinate-based MLP. 

\noindent 
\textbf{Surface Normals} Gradients of the SDF at the zero level set point $\mathcal{S}$ towards the surface normals $\mathcal{S}$, 
\begin{align}
    \mathbf{n}(\textbf{x}) = \dfrac{\nabla_{\mathbf{x}} f_{\mathcal{S}}(\mathbf{x})}{\|\nabla_{\mathbf{x}} f_{\mathcal{S}}(\mathbf{x})\|} \quad \mathbf{x} \in \mathcal{S} 
    \label{eq:sdf_normals}
\end{align}
%By learning signed distance fields, our method is able to generate smooth estimates of object geometry by providing a helpful inductive bias for rendering, and the decomposition of geometry and diffuse radiance. 

\noindent 
\textbf{Surface Curvature} We employ differential geometry techniques developed by Novello \textit{et al.} \cite{novello:i3d:2022} to estimate curvature for neural implicit surfaces. In particular, we estimate the mean curvature  $K(\mathbf{x})$ for the implicit surface from the divergence, $\boldsymbol{\nabla} $ of the surface normals
\begin{align}
    K(\mathbf{x}) = \frac{\boldsymbol{\nabla} \cdot \mathbf{n}(\mathbf{x})}{2} 
    \label{eq:sdf_curvature}
\end{align}
Mean curvature approximates the surface with an osculating sphere. Our approach also works for more generalized notions of curvature through the shape operator, at the cost of higher computational complexity. We refer our readers to the supplement for the general case. 
% \AD{Akshat Talk about mean curvature}

\begin{figure}
    \centering
    \includegraphics[width=\columnwidth]{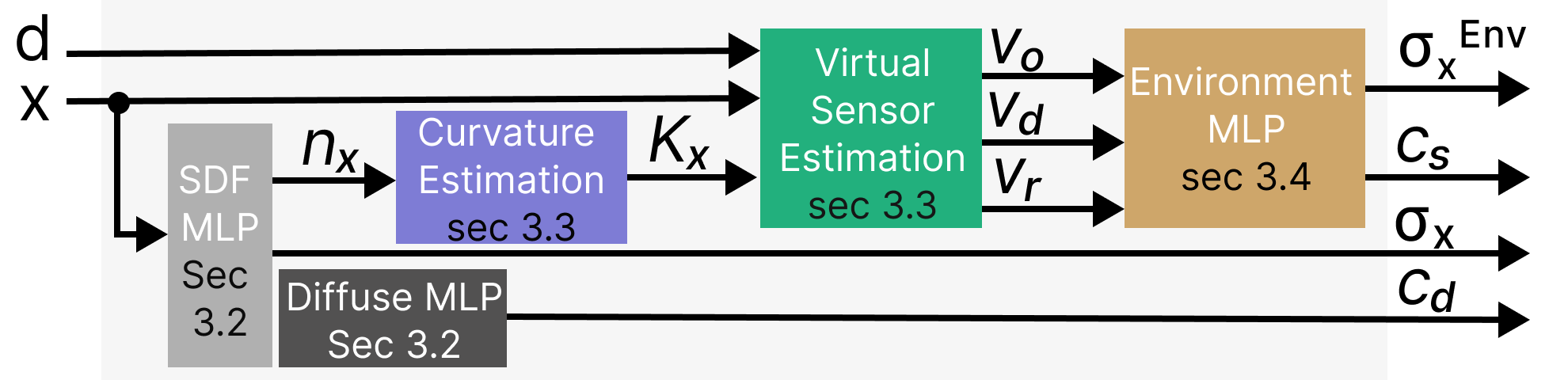}
    \caption{\textbf{Overview of our proposed architecture}.}
    \label{fig:architecture}
\end{figure} 
\noindent
\textbf{Diffuse Radiance} We sepreate the captured radiance at the obsever camera with diffuse radiance, that depends on the glossy object's albedo, and specular radiance that depends on the environment radiance. The diffuse radiance does not have any view dependance and only depends on surface point $\mathbf{x}$. We denote the diffuse radiance as $f_{d}$ model it using a coordinate-based MLP (Fig. \ref{fig:architecture}).  

\noindent
\textbf{Volume Rendering}
 As proposed in \cite{yariv2021volume}, we perform volumetric rendering on the SDF. We define the volume density $\sigma(\mathbf{x})$ as the cumulative distribution function (CDF), denoted as $\Psi(s)$, applied to $f_{\mathcal{S}}$: 
% \begin{gather*} 
\begin{align} 
\label{eq:sdf_to_sigma}
\sigma(\mathbf{x}) = \alpha \Psi_{\beta}(f_{\mathcal{S}})
% \end{gather*} 
\end{align} 
In contrast to \cite{yariv2021volume}, however, we only aim to recover the diffuse radiance of the object along a particular ray. We define a function $f_{d}$ that estimates the diffuse radiance at each point, $\mathbf{x}$, along the ray. To get the final diffuse radiance along a given primary ray, $\mathbf{r}_p(t)$, we perform volumetric rendering: 
% \begin{gather*} 
\begin{align} 
\label{eq:vol_rend}
\hat{\mathbf{c}}_{d}(\mathbf{r}) = \int_{0}^{\infty} f_{d}(\mathbf{r}(t), f^k_{\mathcal{S}}(\mathbf{r}(t))\tau(t) dt
% \end{gather*} 
\end{align} 

\noindent Note that there is no view dependence in  Eq. \ref{eq:vol_rend} and intermediate features, $f^k_{\mathcal{S}}$, are used as input. $\tau(t)$ is the accumulated transmittance along the ray. 

% and the estimated surface normals $\hat{\mathbf{n}}(t) = \frac{\nabla_t f_{\mathcal{S}}(\mathbf{r}(t))}{|| \nabla_t f_{\mathcal{S}}(\mathbf{r}(t)) ||}$ thus only modeling the diffuse radiance of the object. $\tau(t)$ determines the accumulated transmittance along the ray.  

\subsection{Objects Surface as Virtual Sensor}
Each pixel, $\mathbf{p}$, with a finite surface area, $\mathbf{dA}_p$, on the real-camera sensor views the surface of the object through a frustum originating at that pixel. The object then samples the environment radiance field through this finite surface converting the finite surface into a virtual pixel with surface area, $\mathbf{dS}$. Through this model, we can interpret the object surface as a virtual sensor consisting of many virtual pixels that sample radiance from the environment field based on geometry of the object and observer viewing direction. We now formulate a virtual pixel based on real camera post and implicit surface geometry. Please refer to Fig. \ref{fig:virtual_cone_with_surface} for a visualzation of the virtual sensor.

% viewed by a pixel as a virtual `sensor' with a pixel area $\mathbf{dS}$ corresponding to real pixel with area $\mathbf{dA}_p$. 

% The object surface is modeled as a virtual sensor that consists of virtual pixels that 

% We show that for each pixel, $\mathbf{p}$, this viewpoint is a virtual cone originating from the surface.  

\label{sec:Virtual_Cones}
% A primary ray, $\mathbf{r_p}(t) = \mathbf{o} + t \mathbf{d}$, centered at pixel $\mathbf{p}_{i,j}$ with finite area, $\mathbf{dA}_p$, intersects the object projecting a finite surface area, $\mathbf{dS}$, as a function of object geometry and viewing direction, $\mathbf{d}$. Given that the reflection is specular, $\mathbf{dS}$ forms the imaging plane, or a single-pixel virtual sensor, that lies on the object plane. This plane serves as the reflecting surface responsible for the accumulated specular radiance at pixel $\mathbf{p}_{i,j}$. We now discuss our method to efficiently and accurately compute $\mathbf{dS}$ using mean curvature. 
Consider a real camera origin as $\mathbf{o}$ and a pixel on the real sensor $\mathbf{p}_{i,j}$ that corresponds to ray direction $\mathbf{d}$. The primary ray for pixel $\mathbf{p}_{i,j}$ is parameterized with ray length $t$ as $\mathbf{r_p}(t) = \mathbf{o} + t \mathbf{d}$
\begin{figure}
    % \centering
    \includegraphics[width=\columnwidth]{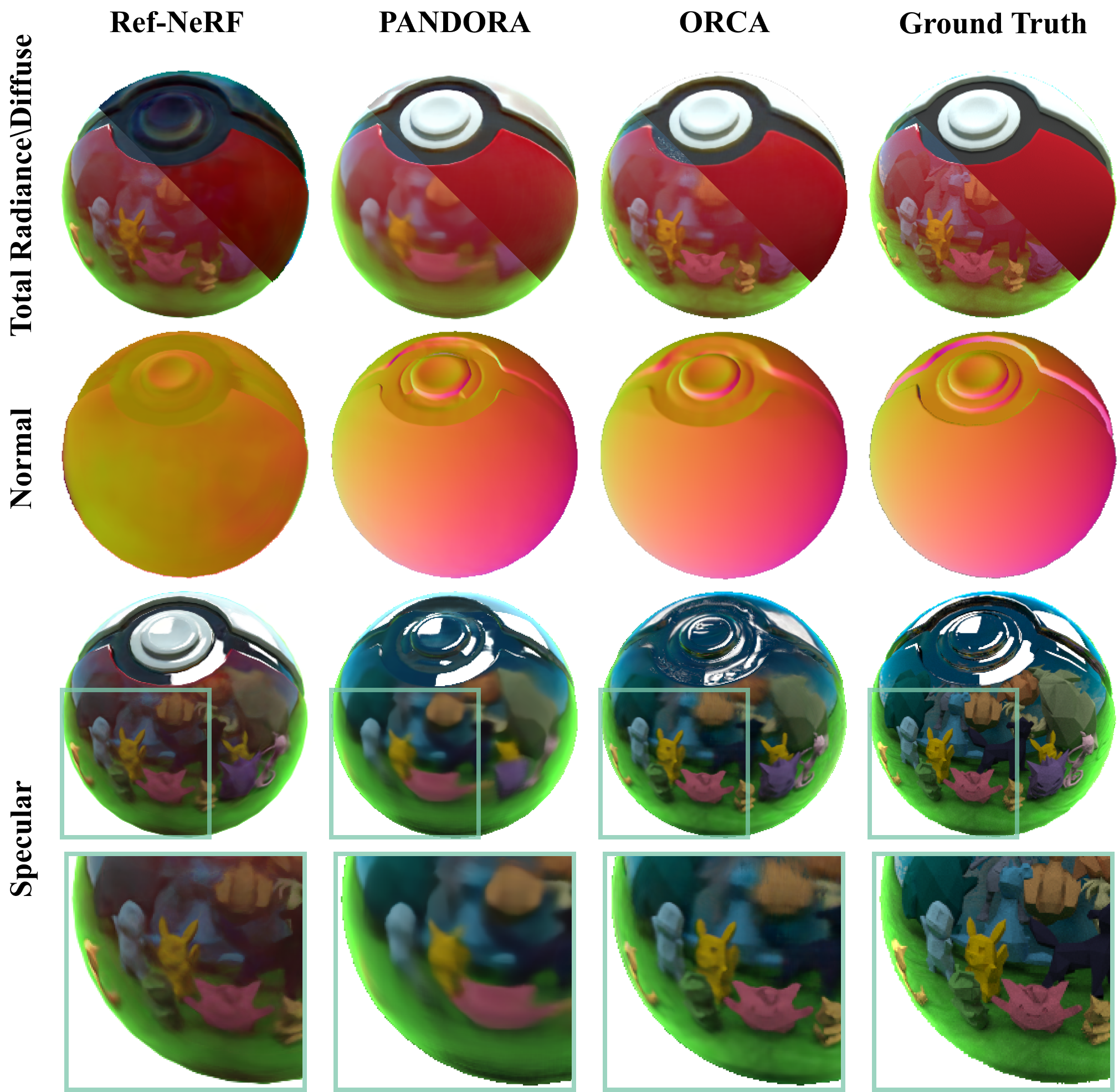}
    \caption{\textbf{Qualitative comparisons of diffuse-specular separation and geometry estimation on rendered dataset}. The environment contains nearby objects with complex occlusions when seen through reflections on the glossy object. RefNeRF fails to perform accurate diffuse-specular separation and PANDORA blurs the nearby objects in the specular map. ORCA can model the complex specular reflections through  environment radiance field.}
    \label{fig:renddataimresults}
\end{figure}

\noindent
\textbf{Casting Real Cones} 
We can approximate the outgoing conical frustum from pixel $\mathbf{p}_{i,j}$ as a cone originating at $\mathbf{o}$ with axis-of-direction $\mathbf{d}$ and radius $\dot{r}$, equivalent to half the distance of the pixel in the x and y directions.
% Our cone approximation is well suited for efficient cone-object intersection computations and forms a good estimate for $\mathbf{dS}$. 
We represent the real-cone as parametric volume,
\begin{align}
\label{eq:ray_cone_eq}
\mathbf{r}_{cone}(\dot{r}, s, \theta) = \dot{r} s \cos(\theta)\hat{\mathbf{e}}_u + \dot{r} s \sin(\theta)\hat{\mathbf{e}}_v + \dot{r} s \mathbf{d}, 
\end{align}
where $\hat{\mathbf{e}}_u$ and $\hat{\mathbf{e}}_v$ are basis vectors in the plane perpendicular to $\mathbf{d}$, $\theta \in [0, \pi]$ and $s \in [0, t_{max}]$
\noindent
\textbf{Virtual Pixel} Virtual pixels are characterized by the intersection of the real cone with the object surface. In Sec. \ref{sec:NIS}, we model local surface properties using mean curvature which enable efficient analytical computations for the virtual pixel parameters even though our approach works with general shape operators. For a sampled point $t_i$ along the ray, we have the surface normals $\mathbf{n}(t_i)$ from Eq. \ref{eq:sdf_normals} and estimated mean curvature $K(t_i)$ from Eq. \ref{eq:sdf_curvature}. The local object surface at $t_i$, can be approximated with an osculating sphere, $\mathcal{O}(t_i)$, centered at $\hat{\mathbf{o_{S}}}(t_i)$ with radius, $R(t_i)$ as follows: 

\begin{gather*} 
\label{eq:smax_eq}
R(t_i) = \frac{2}{K_{t_i}} \\
\hat{\mathbf{o_{S}}}(t_i) = \mathbf{r}_p(t_i) + R(t_i) \cdot \hat{\mathbf{n}}({t_i})
\end{gather*} 
Note that for concave surfaces $K_{t_i} < 0$, so $\hat{\mathbf{o_{S}}}$ will lie outside the object and, for $K_{t_i} > 0$, $\hat{\mathbf{o_{S}}}$ will lie inside the object. 
% Finally, we have a second-order approximation for the object surface $\hat{\mathcal{S}}_{t_i}$: 

% % \begin{gather*} 
% \begin{align} 
% \label{eq:smax_eq}
% \hat{\mathcal{S}}_{t_i} \thickapprox \bigg ( (\mathbf{x} - \hat{\mathbf{o_{S}}})^2 = \big ( \frac{2}{ \mathrm{div} \hspace{1mm}\hat{\mathbf{n}}({t_i}) } \big ) ^2 
%  \bigg )
% % \end{gather*} 
% \end{align} 
The edges of the virtual pixel for $\mathbf{r}_{\mathbf{p}}(t_i)$ would lie at the intersection of the osculating sphere $\mathcal{O}(t_i)$ and the primary cone given by $\mathbf{r}_{cone}$. Computing exact cone-sphere intersections are computationally expensive so we approximate the cone-sphere intersection using rays bound cone-sphere interectional surface $\mathbf{dS}$. We consider four rays that bound the cone and sample them at $\theta_j \in \{ 0, \pi/2, \pi, 3\pi/2\}$ with Eq. \ref{eq:ray_cone_eq}. We perform intersections of the corresponding bounding rays with the osculating sphere $\mathcal{O}(t_i)$ to get corners of the virtual pixel $\mathbf{ds}_j$. These ray sphere intersections can be computed analytically in an efficient manner. 

% We now have a formulation that bounds the maximum surface area $\mathbf{dA}_{t_i}$ visible to the pixel, $\gamma(t_i, \dot{r}, \theta)$, in addition to an efficient approximation for local object geometry, $\hat{\mathcal{S}}_{t_i}$. Thus, we can now approximate the maximum surface area on the object visible to the pixel at $t_i$, $\mathbf{dS}_{t_i}^{max}$ using ray-sphere intersection. We first compute ray-osculating-sphere intersection points, denoted by ${I}_{t_i}(\cdot)$, with the following equation: 

% \begin{gather*} 
% \begin{align} 
% \label{eq:intersection_pts}
% \mathbf{I}(t_i, \dot{r}, \hat{\mathbf{o_S}}, R, \theta) = \bigg \{ \big \| \gamma(t_i, \dot{r}, \theta) - \hat{\mathbf{o_{S}}}_{t_i} \big \|^2 - R_{t_i}^2 \bigg \}
% \end{align} 
% % \end{gather*} 

\noindent
\noindent
\textbf{Virtual Cone Origin}
With an estimate of the virtual pixel surface area, we can now compute the virtual cone that samples the environment.  
% In the following section, we discuss our formulation to estimate the virtual center-of-projection, $\mathbf{o_v}$. Our method uses local estimates of the surface, $\hat{\mathcal{S}}_{t_i}$, and computes virtual viewpoints by solving a closed-form solution for linear least squares. We empirically show that our estimated virtual viewpoints lie on the caustic surface of the object and serve as a good approximation \ref{fig:pipeline}. In the equations below, we have omitted $t_i$ for clarity, but all the calculations are performed along all points sampled on the primary ray. 
We first compute normal vectors at virtual pixel corners $\mathbf{ds_j}$ from the center of osculating sphere $\hat{\mathbf{o}_{\mathbf{S}}}$%found in Eq. \ref{eq:intersection_pts}: 
% \begin{gather*} 
\begin{align} 
\label{eq:normals_pts}
\hat{\mathbf{n_j}} = \frac{\mathbf{ds_j - \hat{\mathbf{o_{S}}} }}{ || \mathbf{ds_j} - \hat{\mathbf{o_{S}}} || }
\end{align} 
% \end{gather*} 

\noindent At each virtual pixel corner, we compute the reflected ray directions, $\omega^r_{j}$, by computing the dot product between the incoming ray directions, $\omega^i_{k}$, and the normals, $\hat{\mathbf{n}}_{k}$, where $\omega^r_{0}$ is the primary ray's reflected vector. 

% As Figure \ref{fig:fig_me} shows, we now have 5 intersection points, $\{\mathbf{k}_i\}_{i=0}^{4} \in \mathbb{R}^3$, on the surface, with $\mathbf{k}_0$ equal to the sampled point on the primary ray. The incoming ray directions at those points, $\omega^i_{k}$ and their associated normal vectors at those intersections, $\hat{\mathbf{n}}_{k}$. We now find the associated reflected vectors, $\omega^r_{k}$:
% \begin{gather*} 
\begin{align} 
\label{eq:reflected_osculate}
\omega^r_{0} = \mathbf{d} - \big (\mathbf{d} \cdot \hat{\mathbf{n}}(t_i) \big ) \hat{\mathbf{n}}(t_i) \\ 
\omega^r_{j} = \mathbf{d}_j - \big (\mathbf{d}_j \cdot \hat{\mathbf{n}}_{j}(k) \big ) \hat{\mathbf{n}}_{j}(k)
% ray_d - 2 * dot(rays_d, normals) * normals
\end{align} 
% \end{gather*} 

$\mathbf{d}_j$ are the incident directions to the virtual pixel corners $\mathbf{ds}_j$. The virtual cone origin is the intersection of these reflected rays at the pixel corners and pixel center. However, these rays might not intersect at a single point so we approximate a virtual origin to be the point that minimizes the sum of distances to the reflected rays $\omega_j$. 
\begin{equation}
    \mathbf{v_o} = \text{argmin}_{\mathbf{v}} \sum_{j}{|(\mathbf{v}-\mathbf{ds_j})\times \omega^r_j|}
\end{equation}
We pose this as a linear least squares problem and compute the psuedo-inverse to efficiently compute the virtual cone origin.
\noindent
\textbf{Virtual Cones Direction.} 
The reflected ray at the center of the virtual pixel reflects the object surface along the direction $\omega_0^r$ from Eq. \ref{eq:reflected_osculate}. We consider this as the direction-of-axis of the virtual cone.
\begin{align} 
\label{eq:virtual_dirs}
\hat{\mathbf{v}}_d = \omega^r_{0} 
\end{align} 
% Our estimated $\hat{\mathbf{v}}_o$ forms the center of projection for the virtual sensor, $\mathbf{dS}$, and we can use this to image the world through $\mathbf{dS}$. We note that $\mathbf{dS}$ images the world and acts like a single pixel sensor located on the object surface. Following the pixel-as-cone approximation used in Section \ref{sec:NIS}, we also approximate this virtual sensor as a virtual cone. Specifically, we cast a virtual cone through some arbitrary surface $\mathbf{dS}$ from $\hat{\mathbf{v}}_o$. We approximate the arbitrary surface $\mathbf{dS}$ with a circle thereby simplifying our calculations. The direction of the virtual cone is given by the principal reflection vector. Alternatively, the virtual cone direction can also be approximated by the direction vector between primary-ray intersection points and center of the osculating sphere. In practice, we use $\hat{\mathbf{v}}_d$ as the latter leads to non-smooth surface normals. 

% \begin{gather*} 
% \begin{align} 
% \label{eq:virtual_dirs}
% \hat{\mathbf{v}}_d = \omega^r_{0} \\ 
% \hat{\mathbf{v}}_d^{\mathbf{O_s}} = \frac{\mathbf{k}_0 - \hat{\mathbf{v}}_o }{|| \mathbf{k}_0 - \hat{\mathbf{v}}_o ||}
% \end{align} 
% \end{gather*} 
\noindent
\textbf{Virtual Cone Radius.} We compute the radius of the cone by treating the  reflection vectors of the bounding rays as the neighboring "pixel" directions. Similar to \cite{barron2021mipnerf}, we can compute the distance between $\{\omega^r_{k_\theta}\}_{\theta=0}^{2\pi}$ and the primary reflected ray $\omega^r_{0}$ in the $(x,y)$ components (omitted below for clarity).

\begin{align} 
% \begin{gather*} 
\label{eq:virtual_radii}
% \dot{r}_v = \frac{1}{4}\sum_{j=1}^4 {| \mathbf{k}_j - \mathbf{k}_0 | }
\hat{\mathbf{v}}_{\dot{r}} = \| \{\omega^r_{k_\theta}\}_{\theta=0}^{2\pi} - \omega^r_{0} \| %\frac{1}{4}\sum_{j=1}^4 {| \mathbf{k}_j - \mathbf{k}_0 | }
\end{align} 
% \end{gather*} 

Finally, for each sampled point $t_i$, we can characterize our single-pixel virtual sensor located at the object surface $\mathbf{dS}$ as a virtual cone with $\hat{\mathbf{v}}_o$ as its apex, $\hat{\mathbf{v}}_d$ as axis-direction, $\hat{\mathbf{v}}_{\dot{r}}$ as the radius. 

% \begin{itemize}
%    \item Primary cone oscullating surface intersection 
%    \item Virtual cone apex
%    \item Virtual cone direction
%    \item Virtual cone radius
% \end{itemize}
\noindent \textbf{Connections to caustics.} Our work takes inspiration from Catadiopritc Imaging systems. To covert objects into cameras, we esentially compute the surface and find a corresponding center-of-projection for this surface-as-sensor. However, unlike conventional perspective cameras, objects don't have a fixed center-of-projection, other than in a few special configurations \cite{Baker2004ATO}, but a locus of viewpoints that vary with object geometry and viewing direction. These viewpoins lie on the "caustic surface" of the object. While typical works in catadioptric imaging use an analytical equation for the caustic surface by assuming known geometry \cite{786912} \cite{937581}, or making assumptions about placement of the observer \cite{axialcones}, our formulation approximates the caustic surface of unknonw geometry through intersection of reflected rays on virtual pixels. We emperically show in supplementary that as the surface area of the virtual pixel goes to 0, $d\mathbf{S} \rightarrow 0$, our method estimates the true caustic of object without assuming geometry. Our method also has applications in estimating the caustic surface of the unknown geometry. 

% \AD{Add if space and time} 
% In order to image the world through the virtual sensor, $\mathbf{dS}_{t_i}$, the sensor must have a virtual center of projection.

\subsection{Environment Radiance Fields}
\label{sec:env_rad_fields}
Our goal is to capture a 5D environment radiance field of the scene by imaging the world through these single-pixel virtual sensors located at the object's surface. We use our formulation of virtual cones to recover 5D environment radiance fields. We define an environment radiance field as $f_{\mathcal{E}}: (\hat{\mathbf{v}}_o, \hat{\mathbf{v}}_d) \rightarrow  (\sigma^{Env}, c_s)$,  
% \begin{align} 
% % \begin{gather*} 
% \label{eq:env_rad_field}
% f_{\mathcal{E}}: (\hat{\mathbf{v}}_o, \hat{\mathbf{v}}_d) \rightarrow  (\sigma^{Env}, c_s)
% \end{align} 
% \end{gather*} 

where $f_{\mathcal{E}}$ outputs opacity and radiance along sampled virtual cones. We note that this view dependent radiance is equivalent to the specular radiance at point $t_i$ sampled along the primary-camera ray $\mathbf{r_p}(t)$. We can render the final specular radiance at pixel $\mathbf{p}_{i,j}$ as follows: 
% \begin{align} 
\begin{gather*} 
\label{eq:vol_rend_1}
\hat{\mathbf{c}}_{s}(\mathbf{r}) = \int_{0}^{\infty} f_{\mathcal{E}}(\hat{\mathbf{v}}_o, \hat{\mathbf{v}}_d) \tau(t) dt \\
\hat{\mathbf{c}} = \hat{\mathbf{c_d}} + \hat{\mathbf{c_s}}
% \end{align} 
\end{gather*} 

Intuitively, $f_{\mathcal{E}}$ learns the 5D radiance field by sampling single-pixel virtual sensors from the object surface area, and must learn geometry and environment radiance that is consistent with multiple views from the object's reflections. Moreover, we can query $f_{\mathcal{E}}$ to render novel viewpoints and associated depths that are beyond field-of-view of the real camera. We volume render each virtual cones by dividing them into conical frustums using Integrated-Positional Encoding as proposed in MipNeRF \cite{barron2021mipnerf}. Our formulation of virtual cones works well with Mip-Nerf's rays-as-cones method. 

\section{Experiments}

Our experiments study the ability of our method to recover 5D environment radiance fields (assessed through quality of predicted surface normals, diffuse radiance, specular radiance, and 3D environment maps) from objects of varying complexity, both in simulation (Fig. \ref{fig:renddataimresults}) and the real-world (Fig. \ref{fig:realdataimresults}). Quantative results are provided in Table [1]. As in prior works on novel view synthesis, we report PSNR and SSIM to evaluate estimated diffuse, specular, and mixed radiance, and report mean angular error (MAE) to evaluate estimated surface normals.

\subsection{Implementation Details}
As in PANDORA, we parameterize $f_{\mathcal{S}}$ with an 8-layer MLP to estimate the surface, and, as in MipNeRF, $f_d$ with 4-layer MLP with input geometric features of size 512 from $f_{\mathcal{S}}$. We follow the sdf-to-opacity conversion and the iterative sampling of the ray proposed in \cite{yariv2021volume}. To aid the network to learn the geometry quickly, we also train $f_{\mathcal{S}}$ with a mask-net as proposed in \cite{dave2022pandora}. We use five losses in our architecture: photometric loss, mask loss \cite{dave2022pandora}, normal loss \cite{verbin2022refnerf}, eikonal loss \cite{yariv2021volume}, and distortion loss \cite{barron2022mip}. Additional training details are discussed in the supplementary materials.

\begin{table}
\resizebox{\columnwidth}{!}{%

\begin{tabular}{|l|l|cc|cc|cc|c|}
\hline
 &  & \multicolumn{2}{c|}{Diffuse Radiance} & \multicolumn{2}{c|}{Specular Radiance} & \multicolumn{2}{c|}{Mixed Radiance} & Normals \\
Scene & Approach & PSNR  & SSIM & PSNR & SSIM & PSNR & SSIM & MAE \\
 &  & $\uparrow$ (dB) & $\uparrow$ & $\uparrow$ (dB) & $\uparrow$ & $\uparrow$ (dB) & $\uparrow$ & $\downarrow$ (\degree) \\ \hline
      & Ref-NeRF & \textbf{17.59} & \textbf{0.7217} & 14.88 & 0.4750 & \textbf{19.58} & \textbf{0.7956} & 62.45 \\
D1 & PANDORA & 13.23 & 0.4759 & 15.12 & \textbf{0.5231} & 12.87 & 0.4607 & 2.387 \\
      & ORCA & 13.29 & 0.4683 & \textbf{16.64} & 0.5148 & 18.23 & 0.5745 & \textbf{1.873} \\ \hline
 & Ref-NeRF & 11.86 & 0.6090 & 15.28 & \textbf{0.7059} & 21.80 & \textbf{0.8643} & 33.92 \\
D2 & PANDORA & 22.53 & 0.8689 & 17.76 & 0.6326 & \textbf{22.73} & 0.7787 & 3.693 \\
 & ORCA & \textbf{23.47} & \textbf{0.8954} & \textbf{18.98} & 0.6954 & 22.31 & 0.8107 &  \textbf{3.568} \\ \hline
\end{tabular}
}
\label{table:allmetrics}
\caption{\textbf{Quantitative evaluation of rendered scenes}. We compare ORCa to other neural rendering techniques that model reflections, including Ref-NeRF and PANDORA, on the globe (D1) and Pokemon (D2) datasets. ORCa is competitive with the comparison methods in accurate diffuse and specular separation, and provides consistent improvement in geometry and specular radiance estimation. }
\end{table}

\subsection{Datasets}

We conduct experiments on both simulated and real-world datasets. Simulated datasets are rendered in Mitsuba2 \cite{nimier2019mitsuba}. Simulated datasets contain a range of increasingly complex object geometries (elephant, Pokeball, and orca) and scenes (living room and Pokemon). We train with 200 views for simulated datasets. We also show results for a real-world dataset \cite{dave2022pandora} capturing a glossy cup with a black vase sitting atop it using 35 views. All datasets will be publically released upon publication.
% Real-world datasets from PANDORA  are used to validate the real-world applicability of our method. More real-world dataset results are included in the supplementary material. All datasets will be publically released. 
% We conduct experiments on both simulated and real-world datasets. Simulated datasets are rendered in Mitsuba2 \cite{nimier2019mitsuba}, which allows us to control material properties, camera angles, illumination, and polarization. Simulated datasets contain a range of object geometries (elephant, pokeball, and orca) and scenes (living room and pokemon). We train with 200 views for simulated datasets. Real-world datasets from PANDORA \cite{dave2022pandora} are also used to validate the real-world applicability of our method. We show results for a real-world dataset capturing a glossy cup with a black vase sitting atop it. We train with 35 views for real-world datasets. More real-world dataset results are included in the supplementary material. All datasets will be publically released upon publication. 

%Polarization data is also captured for the real-world datasets using a polarization lens. We note that polarization data is only used in our polarization ablation and our method can be used either with or without it. \TK{Move previous sentence to supplementary once we move polarization fig.}

\subsection{Comparisons with Baselines}

\begin{figure}
    \centering
    \includegraphics[width=\columnwidth]{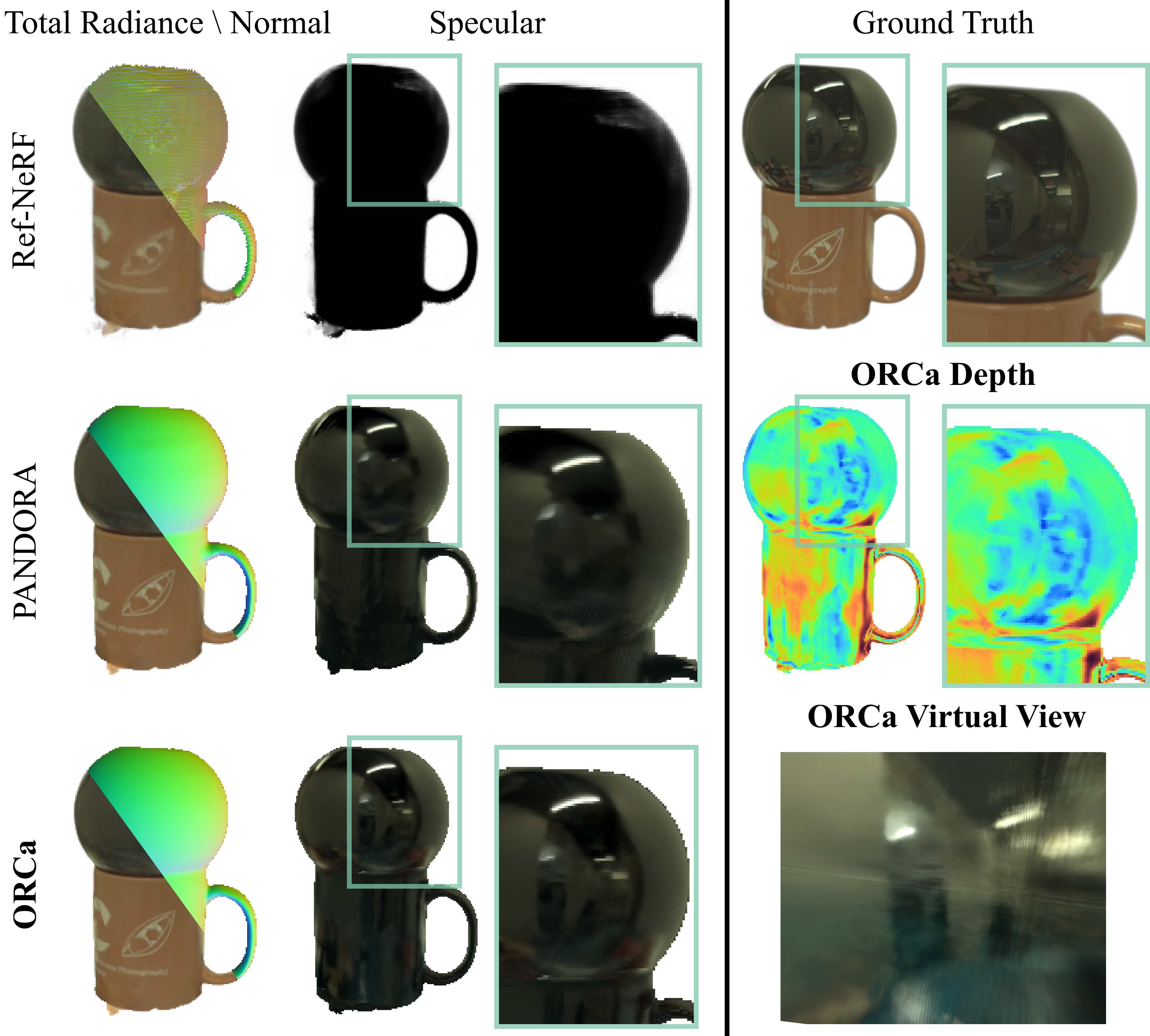}
    \caption{\textbf{Comparisons on real dataset}. Using a real-world dataset with only 35 views, ORCa can model the sharp specularities on the ball arriving from regions of the nearby scene, such as the table, and far away scene regions, such as the hallway, to learn an environment radiance field. We query the radiance field for depth of the far hallway (blue) and the nearby objects, such as the table (red). We also render novel viewpoints that are beyond the field-of-view of the observer camera and show that ORCa interpolates well between those views.}
    \label{fig:realdataimresults}
\end{figure}

We compare our method to other neural rendering techniques that model reflections, Ref-NeRF and PANDORA.

% \textbf{Ref-NeRF \cite{verbin2022refnerf}}. Ref-NeRF provides a method for measuring outgoing radiance on glossy objects by reflecting the viewing vector about the estimated normal, significantly improving upon prior NeRF methods. Since Ref-NeRF focuses on improving novel view synthesis on scenes containing glossy objects, their method reflects rays and recovers a 2D radiance field, whereas we shoot virtual cones from the glossy object and focus on recovering the environment map. Unlike our method, Ref-NeRF does not recover any 2D or 3D environment maps. 

% \textbf{PANDORA \cite{dave2022pandora}}: PANDORA proposes a method to decompose images into specular and diffuse radiance, using either multi-view images or multi-view images and polarization. PANDORA is able to generate 2D environment maps, but not 3D. We use PANDORA without polarization when comparing to our method.

% \begin{itemize}
%     \item \textbf{Ref-NeRF \cite{verbin2022refnerf}}: 
%     %recovering total radiance and surface normals of glossy objects from multi-view RGB images by modeling rays that are reflected about estimated normals. Unlike our method, Ref-NeRF does not recover diffuse radiance, specular radiance, or any 2D or 3D environment map of the scene.
%     \item 
% \end{itemize}

We first discuss results in Fig. \ref{fig:parallax_example} which show the advantages of recovering a 5D radiance field with close-by objects as they often cause occlusions which cannot be modeled by 2D environment maps. By estimating the radiance field, we can image behind occluders through sampling novel viewpoints such as the translated viewpoints shown in Fig. \ref{fig:parallax_example}. Moreover, we can also show depth to surroundings from these vritual viewpoints. We provide additional examples of depth and beyond \textit{field-of-view} novel-view synthsis in the supplementary materials.

While Ref-NeRF and PANDORA learn 2D environment radiance fields, ORCa recovers a 5D environment radiance field. As shown in Fig. \ref{fig:renddataimresults} and \ref{fig:realdataimresults}, ORCa estimates more accurate surface normals than other methods. While the total radiance predicted by Ref-NeRF and PANDORA are visually similar, the surface normals are less smooth than ORCa. We also observe that ORCa is able to achieve better diffuse and specular radiance separation than PANDORA, which is evident in the Pokeball surface normals Fig. \ref{fig:renddataimresults}. In these examples, PANDORA recovers blurry specular radiance. We see that ORCa's predicted depth is highly interpretable and matches the underlying geometry of the environment, as shown in Fig. \ref{fig:parallax_example}. Even on cylinderical real-world datasets, such as the black vase in Fig. \ref{fig:realdataimresults}, the nearby hallway is visible in both the virtual view and depth, despite never being in the field of view of the primary camera. Unlike Ref-NeRF, our primary objective is not to perform novel-view synthesis, but instead to capture the environment radiance field from the object surface.

As shown in Table 1, ORCa is competitive with both Ref-NeRF and PANDORA in estimating diffuse radiance, specular radiance, mixed radiance, and normals. Although the comparison methods slightly outperform ORCa in full, mixed-radiance scene rendering, ORCa consistently provides better specular radiance and object geometry estimation across scenes and viewpoints. This is again indicative of a key strength of ORCa; whereas existing approaches aim to perform novel-view synthesis on reflective objects, ORCa specifically focuses on accurate specular reflection retrieval for environment radiance field modeling. This is achieved through accurate object geometry modeling, which enables high-accuracy specular radiance estimation, thereby aiding in beyond \textit{field-of-view} novel-view synthesis. 

\subsection{Impact of Correct Virtual Cones}
\begin{figure}
    \centering
\includegraphics[width=\columnwidth]{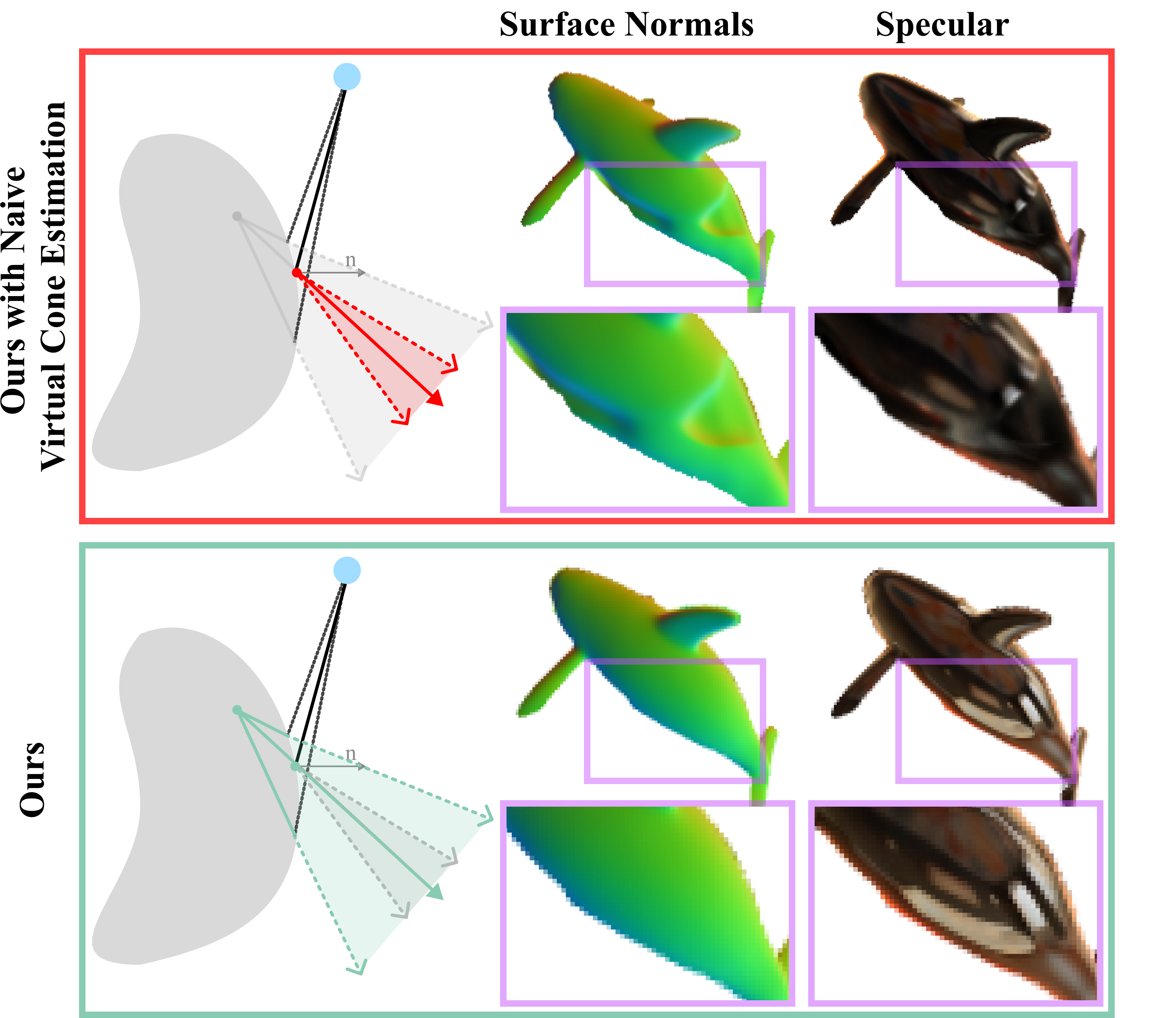}
    \caption{\textbf{Ablation on virtual cone formulation}. We study the impact of our proposed virtual cone formulation for recovering the 5D radiance fields compared to a naive formulation where the emitted cones' origins are placed on the object surface. In contrast, we place the emitted cones' origins within the object based on caustics and compute the cone radius. We see that this leads to smoother estimated surface normals (left) and more accurate estimated specular radiance (right).}
    \label{fig:ivc_vs_cvc}
\end{figure}

We base our method on a physically accurate formulation by modeling ray-cone intersections and using the surface as a virtual sensor, as described in Sec \ref{sec:Virtual_Cones}. Naively, the origins of the virtual cones could instead be placed at the intersection of the primary camera ray and surface, in essence placing a Mip-NeRF at each intersection point. This alternative formulation would not be physically accurate and Fig. \ref{fig:ivc_vs_cvc} shows the improvement that we achieve, underscoring the importance of correctly modeling virtual cones.
% using a naive virtual cone formulation significantly degrades performance, leading to worse surface normals and specular radiance.

%\subsection{Polarization as a Cue}
%Our method can also incorporate polarization data to improve diffuse and specular color separation. As shown in Fig. \ref{fig:polarization}, exploiting polarization cues leads to improved performance, especially on complex geometries, where specular and diffuse separation from only multi-view RGB is more difficult.

%\begin{figure}
%    \centering
%\includegraphics[width=0.5\textwidth]%%{figures/polarization_ablation.png}
%    \caption{\textbf{Polarization Ablation} Our method can be used either with or without polarization data. If polarization data is available, it can easily be incorporated to ORCA, resulting in improved diffuse and specular separation, especially on scenes containing glossy objects with complex geometries. \TK{Current results here are from training views.}}
%    \label{fig:polarization}
%\end{figure}

% \subsection{Our additional applications}

% \begin{itemize}
%     \item Synthetic View Synthesis
%     \item 3D surroundings
%     \item Material editing
%     \item Virtual object insertions
% \end{itemize}

\section{Conclusion}

In conclusion, we present a method to convert glossy objects with unknown geometry and texture into radiance-field cameras that capture the environment radiance field around them. Our method recovers object geometry and diffuse radiance, in addition to capturing the depth and radiance of the object's surroundings from its perspective. Our modeling of environment as a radiance field is effective in recovering close-by objects (Fig \ref{fig:renddataimresults}), in addition to being occlusion aware (Fig \ref{fig:parallax_example}). Moreover, by recovering the environment radiance field we can perform beyond \textit{field-of-view} novel-view synthesis. Our work can unleash applications in virtual object insertion and 3D perception, e.g. inferring information beyond the line-of-sight of the camera using predicted virtual views and depth.

%This work can be used for 3D perception beyond-the-line-of-sight of the camera by processing predicted virtual views and depth. In addition, our method enable virtual object insertion in areas not in the field of view of the camera.

% In conclusion, we present a method that allows glossy objects to be used as virtual cameras. Our method enables both environment maps and depth beyond the field-of-view of the primary camera to be recovered from a virtual camera placed. First, we recover the geometry of the glossy object using multi-view RGB images of the object from the primary camera. Then, by computing cones that originate from the center of the glossy object onto the surrounding scene, we recover the environment radiance field. The key insight is that we can treat the surface of the glossy object as a virtual sensor. Our results show that our method yields accurate reconstructions both of the glossy object and of the surrounding environment. This work can unleash new applications in novel view synthesis, material editing, virtual object insertion, and beyond.

Our formulation of the radiance field beyond the conventional direct-line-of-sight radiance field can enable further areas of research that aim to to extract more information from the environment and the objects present in it. 

% in how much can information does an image really have? We show that we can in fact capture the radiance field of the areas in the scene that is not beyond the field-of-view of the original camera by capturing the radiance field from different objects. This effectively transforms regular objects into radiance field cameras that can provide new perspectives on to the scene. We believe our work opens up new avenues that go estimation of 2D environment maps from objects in addition to  

%%%%%%%%% REFERENCES
{\small
\bibliographystyle{ieee_fullname}
\bibliography{egbib}
}

\end{document}